# Autoencoding with a Learning Classifier System: Initial Results

Larry Bull

Computer Science Research Centre

University of the West of England, Bristol BS16 1QY UK

Larry.Bull@uwe.ac.k

**Abstract**

Autoencoders enable data dimensionality reduction and are a key component of many (deep) learning systems. This short paper introduces a form of Holland's Learning Classifier System (LCS) to perform autoencoding building upon a previously presented form of LCS that utilises unsupervised learning for clustering. Initial results using a neural network representation suggest it is an effective approach to data reduction.

**Introduction**

Learning Classifier Systems (LCS) [Holland, 1976] typically use evolutionary computing techniques to design a set of production rules for reinforcement learning tasks [Holland & Reitman, 1978], although they have also long been used for supervised learning tasks [Bonelli et al., 1990]. A more recent innovation in LCS is the use of unsupervised learning, specifically for clustering [Tammee et al., 2006]. Autoencoding is an unsupervised learning approach to data dimensionality reduction (eg, [Hinton & Zemel, 1994]), typically used within deep learning architectures. Autoencoding via a single neural network has previously been used with LCS (eg, [Matsumoto et al., 2017]]) and deep neural networks have been used to provide actions/outputs [Kim & Cho, 2019].

Somewhat akin to autoencoding, Booker [1988] presented a form of LCS which extends the principle of using an evolutionary algorithm (EA) to discover any underlying regularities in the problem space, dividing the task of learning such structure from that of supplying appropriate actions to receive external reward. Here a separate LCS exists for each of these two aspects. A first LCS receives binary encoded descriptions of the external environment, with the objective to learn appropriate regularities through generalizations over the input space. This is seen as analogous to learning to represent categories of objects. The matching rules not only post their actions/outputs onto their own internal memory/message list but some are passed as inputs to a second LCS. The second LCS receives reward when it correctly exploits such categorizations with respect to the current task (see [Bull & Fogarty, 1994] for a related LCS using only an EA).

This paper presents initial results from using an LCS to perform autoencoding via a neural network rule representation scheme (after [Bull, 2002]). The outputs of which may then be used in conjunction with another LCS, etc.

**YCSAE: Autoencoding with LCS**

YCS [Bull, 2005] is a simple LCS derived from XCS [Wilson, 1995] and has been used to present a simple form for clustering - termed YCSC [Tammee et al., 2006]. YCSC maintains a rulebase of a number (*N*) of rules, associated with each is a scalar which indicates the average error ($\varepsilon$) in the rule's matching process and an estimate of the average size of the niches (match sets - see below) in which that rule participates ($\sigma$). In this paper YCSC is extended for autoencoding (YCSAE) first such that each rule is represented by a multi-layered perceptron. For a task with $I$ inputs, each rule has *I* input nodes, *H* hidden layer nodes, and *I*+1 output layer nodes. Each node is fully connected to the previous layer, contains a bias, and uses the sigmoid transfer function.

On receipt of an input data, each rule/MLP processes the input in the traditional way and a match to the given input is indicated by the "extra" output layer node having an activation level greater than 0.5. Any rule that matches the input is tagged as a member of the current match set [M].

Learning in YCSAE consists of updating the matching error which is derived from the root mean squared error with respect to the input and the corresponding value on each output layer node of a rule in the current [M] using the Widrow-Hoff delta rule with learning rate $\beta$:

$$\varepsilon_j \leftarrow \varepsilon_j + \beta \left( \left( \Sigma (I_i - O_i)^2 \right)^{0.5} - \varepsilon_j \right) \quad (1)$$

Next, the niche size estimate is updated:

$$\sigma_j \leftarrow \sigma_j + \beta ( |[M]| - \sigma_j) \quad (2)$$

YCSAE employs two discovery mechanisms, a niche EA and a covering operator. As in XCS, a time-based mechanism is used under which each rule maintains a time-stamp of the last system cycle upon which it was consider by the EA. The EA is applied within the current niche when the average number of system cycles since the last EA in the set is over a threshold $\theta_{GA}$. If this condition is met, the EA time-stamp of each rule in the niche is set to the current system time, a parent is chosen according to their fitness using standard roulette-wheel selection, and their offspring (potentially) mutated, before being inserted into the rulebase based on niche size estimate.

The EA uses roulette wheel selection to determine two parent rules based on the inverse of their error:

$$f_i = \frac{1}{\varepsilon^{\upsilon} + 1} \qquad (3)$$

Offspring are produced via mutation (probability $\mu$) where a gene (connection weight) is mutated by adding an amount + or - $rand(m_0)$, where $m_0$ is a fixed real, rand picks a real number uniform randomly from $(0.0, m_0]$, and the sign is chosen uniform randomly. Replacement of existing members of the rulebase uses roulette wheel selection based on estimated niche size. If no rules match on a given time step, then a covering operator is used which creates a random rule/network until it matches the current input, which then replaces an existing member of the rulebase in the same way as the EA.

**Results**

Figure 1 shows the average error of the rulebase of neural autoencoders over time for $l = 11$ and $l=20$ binary input data sets sampled randomly, averaged over 10 runs. The input consists of either all 0s or all 1s, with 10% random noise (bit flipping). Here $N=1000$, $H=5$, $\varepsilon_0 = l/2$, $\sigma_0 = N/2$, $\theta_{GA}=25$, $\mu=0.05$, $m_0=0.1$, $\nu=50$, and $\beta=0.2$.

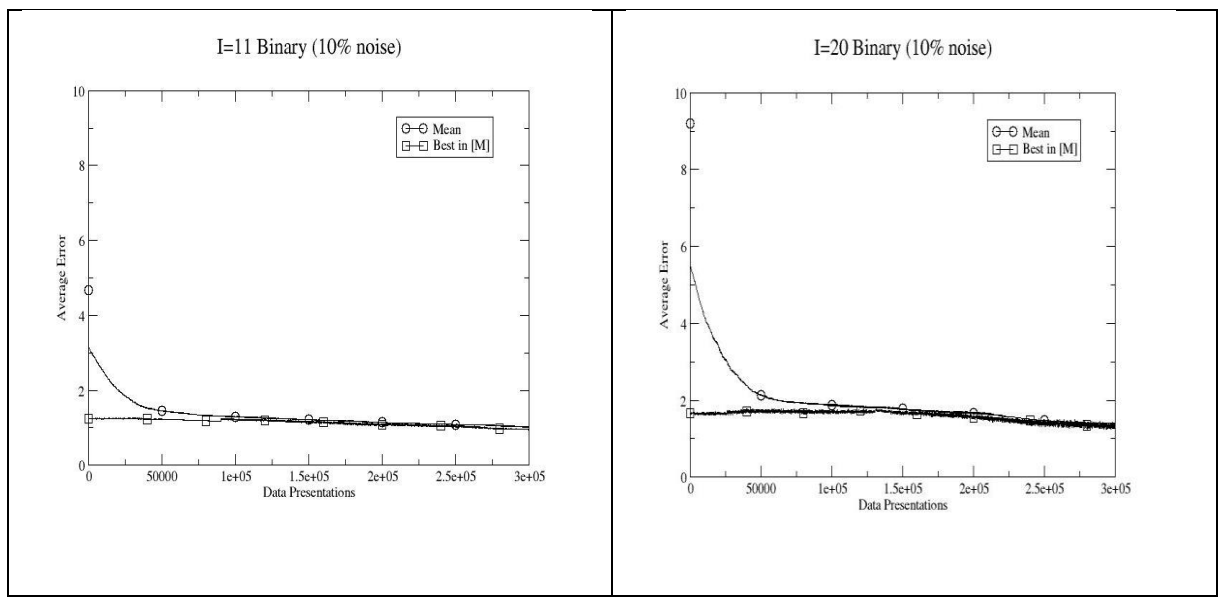

Figure 1: Typical evolution of autoencoder error over time for 11 and 20-bit noisy data.

As can be seen, the average (root mean squared) error decreases to around 1-1.5% in both cases, suggesting effective dimensionality reduction at the hidden layer. Note very little parameter tuning was attempted and typical default values used.

**Conclusion**

Autoencoding is a key component of many deep learning systems and this paper has presented very initial results from using an LCS to perform such dimensionality reduction. LCS offer the potential advantage of building both an ensemble of autoencoders. Moreover, given their basis in EAs, they do not require the existence of helpful gradients within the weight space, although gradient-based search can be added to improve performance (after [O'Hara & Bull, 2007]). Current work is adding a second classification LCS to explore the effectiveness of the suggested dimensionality reduction shown here on various data sets (eg, after [Matsumoto et al., 2017]).